\documentclass[sigconf,natbib=true,anonymous=false]{acmart}
\usepackage{caption}
\usepackage{subcaption}
\usepackage{multicol}
\usepackage{multirow}

\AtBeginDocument{%
  \providecommand\BibTeX{{%
    \normalfont B\kern-0.5em{\scshape i\kern-0.25em b}\kern-0.8em\TeX}}}

\title{PoKE: A Prompt-based Knowledge Eliciting Approach for Event Argument Extraction}
\author{Jiaju Lin}
\affiliation{
  \institution{East China Normal University}
    \city{Shanghai}
    \country{China}}
\email{jiaju\_lin@stu.ecnu.edu.cn}
\author{Qin Chen}
\affiliation{
  \institution{East China Normal University}
  \city{Shanghai}
  \country{China}}

\email{qchen@cs.ecnu.edu.cn}

\begin{abstract}

Eliciting knowledge from pre-trained language models via prompt-based learning has shown great potential in many natural language processing tasks. Whereas, the applications for more complex tasks such as event extraction are less studied, since the design of prompt is not straightforward for the structured event containing various triggers and arguments.
In this paper, we present a novel prompt-based approach, which elicits both the independent and joint knowledge about different events for event argument extraction. The experimental results on the benchmark ACE2005 dataset show the great advantages of our proposed approach. In particular, our approach is superior to the recent advanced methods in both fully-supervised and low-resource scenarios.
\end{abstract}

\begin{document}

\maketitle

\section{Introduction}
\begin{figure}[t]
    \centering
    \includegraphics[width=1\linewidth]{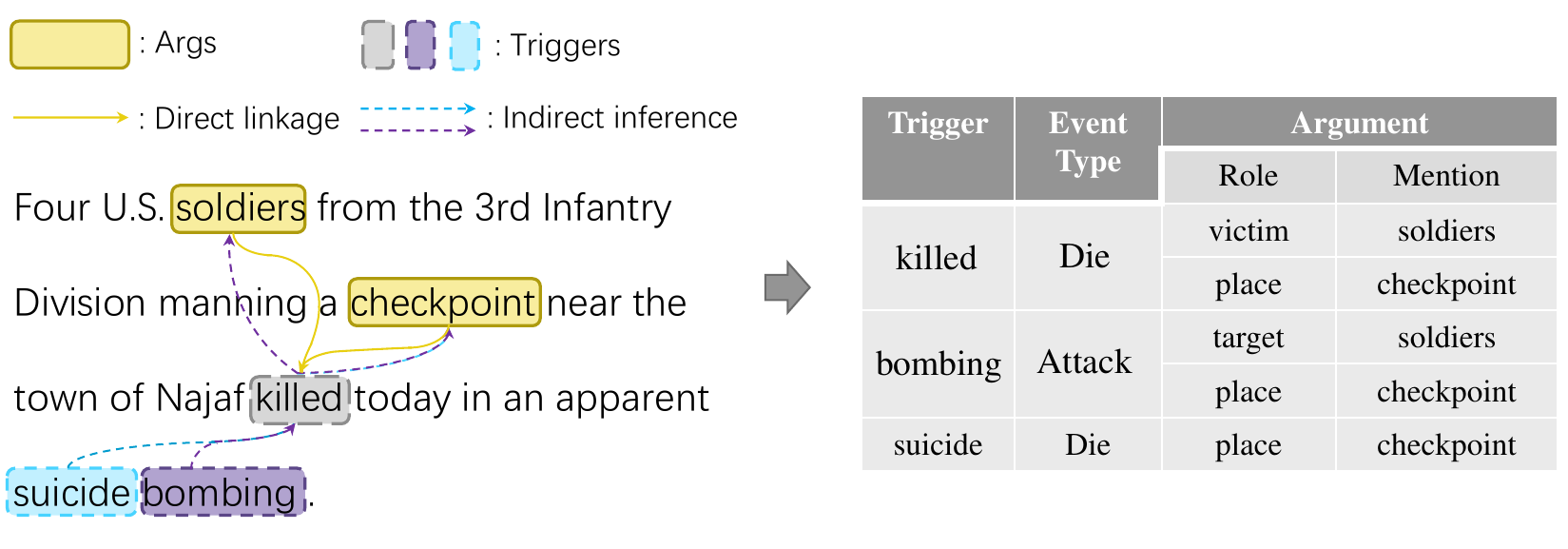}
    \vspace{-2mm}
    \caption{An example of event argument extraction from ACE2005 corpus. There are three events in this sentence. While only one trigger word, `killed', links to the two arguments directly. For other triggers, the arguments need to be inferred via the information provided by the `killed' event. }
    \label{fig:Introduction}
    \vspace{-4mm}
\end{figure}

Event argument extraction (EAE) is a key step towards event extraction \cite{zhou2021role,wang2019hmeae}, which plays an important role in many downstream tasks such as event retrieval and social media analysis \cite{he2021rumor,sarwar2020query}. It identifies  argument spans with a given trigger and event type. 
As shown in Figure \ref{fig:Introduction}, argument mentions of particular roles are identified from the raw sentence for different events.

Recently, there is a trend that utilizes manually designed templates to enhance pre-trained language models (PLMs) for event argument extraction, e.g., the question answering (QA) and prompt learning based methods. \citet{du-cardie-2020-event} and \citet{liu-etal-2020-event} transformed EAE into a QA paradigm, where the question templates were adapted from the annotation guidelines for event extraction. Inspired by the effectiveness of prompt learning in various tasks \cite{li-liang-2021-prefixtuning}, \citet{li-etal-2021-conditionalG} took advantages of the pre-trained language model, namely BART \cite{lewis-etal-2020-bart}, and generated a filled-in template with arguments conditioned on the unfilled template and given context. 
Despite these inspiring progress, current methods still struggle due to the following two issues.

First, the interactions among different arguments are not well studied. \citet{li-2020-Multi-turn} intended to model the argument relations by introducing historical argument embeddings with multi-turn QA. However, the order of questions is determined according to the F1 score on the validation set, which is sub-optimal, especially under few-shot scenarios with limited samples in training and validation. 


The second issue lies in that most existing prompt-based methods rely on an encoder-decoder backbone to generate the argument \cite{li-etal-2021-conditionalG, DBLP:journals/corr/abs-2108-12724}, which does not work well when there are several mentions for an argument role. Specifically, the decoder decodes arguments in an auto-regressive paradigm, resulting in error propagation. For example, if there are more than one victim mentions, such as solders, drivers and so on for the `Die' event, the first error prediction would decrease the precision of following arguments. Besides, the encoder-decoder structure entails more parameters and requires higher training cost.

To address the above issues, we propose a PrOmpt-based Knowledge Eliciting approach (PoKE), which elicits knowledge from PLMs with two prompt strategies, namely \textit{single argument prompt} and \textit{joint argument prompt}. The single argument prompt makes use of the knowledge for each argument independently. In contrast, the joint argument prompt elicits more complementary knowledge about triggers and arguments in the current event and related events, which well captures the intra and inter interactions for enhancing EAE. As shown in Figure \ref{fig:Introduction}, our approach can infer the arguments of `place' and `target' for the `suicide' and `bombing' events with the knowledge of the `Die' event. Moreover, instead of introducing a transformer decoder, we present a Question-Aware Sequence Tagger to obtain all argument spans for an argument role, which works well when there are more than one argument mentions in the context. We evaluate our approach on the EAE benchmark dataset, namely ACE2005. Experimental results show that our approach outperforms the state-of-the-art methods. It is also notable that our approach is capable of better learning in the low-resource scenarios as well as transfer learning.

The contributions can be summarized as follows:(1) We develop various prompt strategies, which can well capture the interactions in intra and inter events for boosting EAE. (2) We explore the effectiveness of integrating prompt-based learning with sequence tagging, which yields better performance with low training cost, especially when there are more than one argument mentions. (3) Extensive experiments on the most widely used benchmark dataset demonstrate the superiority of our approach under both fully supervised and few-shot learning scenarios.

\begin{figure*}[t]
\vspace{-4mm}
\centering
\includegraphics[scale=0.30]{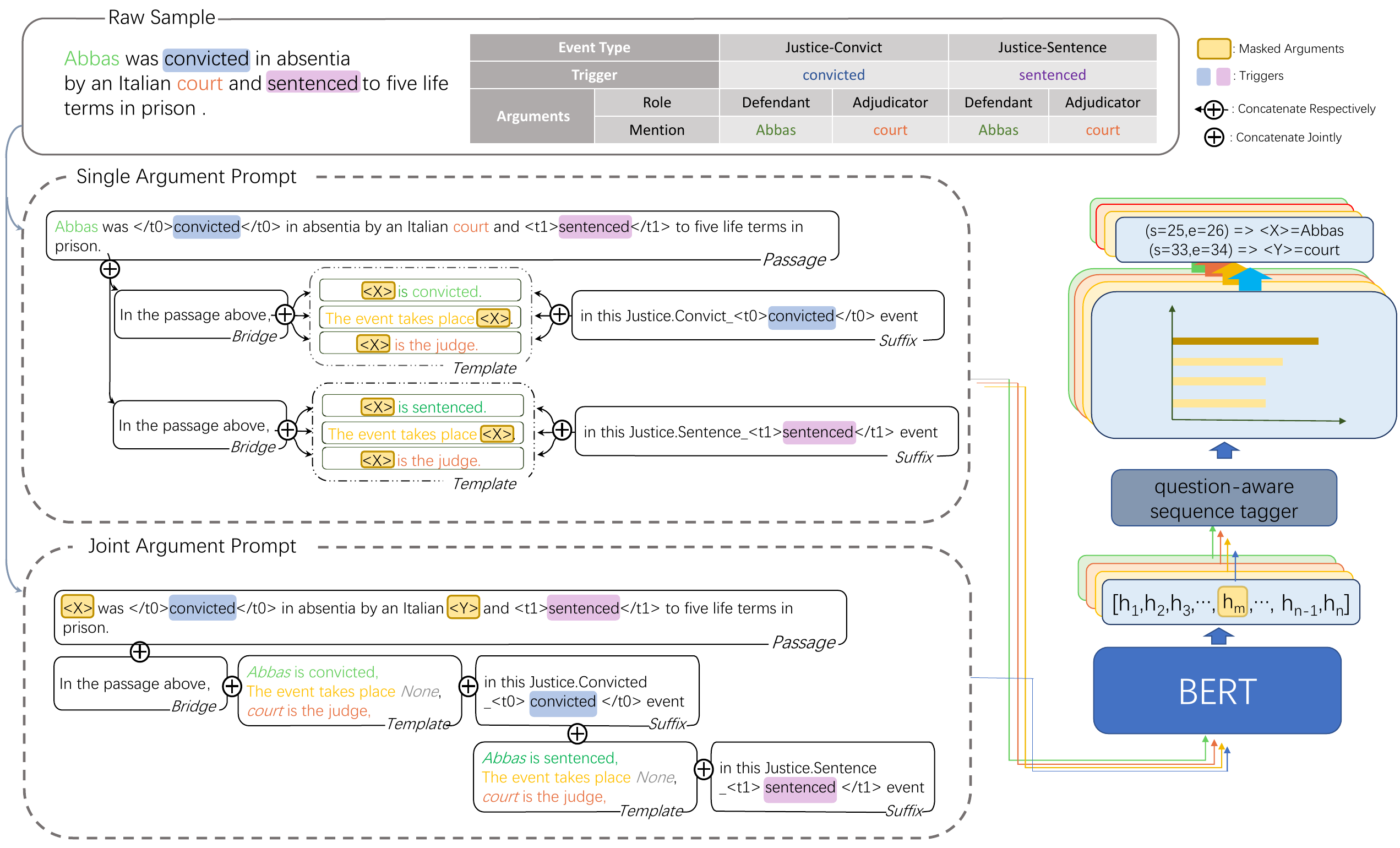}
\vspace{-3mm}
\caption{Architecture of PoKE for event argument extraction. }
\label{fig:arg}
\vspace{-2mm}
\end{figure*}

\section{The Proposed Approach}
\subsection{Framework Overview}
The framework of our approach is shown in Figure \ref{fig:arg}. To be specific, we present two prompt strategies, which wrap the adapted input sentence $x$ with different prompts, by a function $\lambda \left ( \cdot, \cdot  \right ) $ :
\begin{equation}
        X_{p} = \lambda \left ( x,p \right ) = {\rm[CLS]} \ x \ p \  {\rm[SEP]}
\end{equation}
where $p$ denotes the prompt obtained by single argument prompt or joint argument prompt with different `[MASK]' positions, $\rm[CLS]$ and $\rm[SEP]$ indicate the start and end of the sequence. Then, we encode the new input $X_{p}$ by a BERT\cite{devlin-etal-2019-bert} encoder. After encoding, we employ a question-aware sequence tagger to obtain the corresponding argument span, based on the context-aware representations of the `[MASK]' tokens. 

\vspace{-0.5em}
\subsection{Prompt Strategies} 
We develop two types of prompts, both of which consist of three components, namely \textit{bridge}, \textit{template}, and \textit{suffix}. The bridge is a section of words that makes the sequence more fluent. The template contains descriptive sentences rendering definitions for each argument derived from the annotation guideline\footnote{All templates for arguments can be found in our code}, and suffix provides information about triggers and event types.

\textbf{Single Argument Prompt.} For the bridge, we apply a series of fixed words as shown in Figure \ref{fig:arg}. Regarding to the template, we utilize the descriptions in the ACE annotation guidelines for each argument role in an event. Particularly, we mask the argument tokens with a `[MASK]' in the template. In addition, we randomly sample an argument that does not occur in the context and add the description in the template as the negative sample for ease of training. As shown in Figure \ref{fig:arg}, the description as ``\textit{The event takes place <X>}" is not mentioned in the context and is used as a negative sample. Furthermore, it is necessary to have the information of the event type and trigger for argument extraction. Thus, we combine the event type and its corresponding trigger as the suffix, where the trigger word is surrounded by a pair of special tokens as $\rm \langle t\rangle$ $\rm \langle /t\rangle$. An example of the single argument prompt is shown in Figure \ref{fig:arg}. Given a raw sample, we first 
clamp the triggers with special tokens and get $x_s$ as the passage. Then we can generate a single argument prompt denoted as $X_{p_s}$ as follows:
\begin{equation}
\begin{aligned}
     p_s & = \ B\ T^e_a\ S_e \\
    X_{p_s} = \lambda \left (x_s,p_s \right ) & = {\rm[CLS]} \ x_s \ p_s \  {\rm[SEP]}
\end{aligned}
\end{equation}
where $B$ indicates the bridge section, $T^e_a$ represents the template of argument $a$ in event $e$, and $S_e$ is the suffix containing the event type and trigger.


\textbf{Joint Argument Prompt.} When human beings try to understand an event, it is natural to utilize all the information in the context. To simulate such a cognition process, we present a joint argument prompt strategy, which attempts to infer arguments with the information of related arguments in the current events and contextual events.
Figure \ref{fig:arg} shows an example of our joint argument prompt.
For the raw sample, besides clamping triggers with special tokens, we additionally mask all the argument tokens and get $x_j$ as the passage. Then, we append the bridge, template, and suffix of each event consecutively after $x_j$. Specifically, the template of each event contains the descriptions for all argument roles without masking. We also add some out-of-scope argument descriptions as the negative sample. Furthermore, we shuffle the argument descriptions to prevent cheating by spurious correlations, like the order of arguments. The suffix is the same as that in the single argument prompt. More concretely, given a raw sample with $k$ events inside, the joint argument prompt $X_{p_j}$ is generated as follows: 
\begin{equation}
\begin{aligned}
     p_j= & B\ [T^{e_1}_{a_1},...,T^{e_1}_{a_{n_1}}]\ S_{e_1},...,[T^{e_k}_{a_1},...,T^{e_k}_{a_{n_k}}] \ S_{e_k}  \\
     & X_{p_j} = \lambda \left (x_j,p_j \right ) = {\rm[CLS]} \ {x_j} \ p_j \  {\rm[SEP]}
\end{aligned}
\end{equation}
where $n_i$ denotes the number of arguments in event $e_i$. 

\vspace{-1em}
\subsection{Question-Aware Sequence Tagging}
Since multiple argument mentions may exist for a specific argument role, we present a question-aware sequence tagging method to locate all the argument spans, which is different from previous research that only extracts one answer span \cite{ram-etal-2021-shot}. Specifically, we first employ BERT as the encoder to obtain the representations:
\begin{equation}
    h_1,h_2,...,h_n=\rm BERT(X_p)   
\end{equation}

Then, we compute the token-level probability of whether a token is the start or end position of an argument based on the representation of the masked token:
\begin{equation}
\begin{aligned}
    P(s=i|X_p)=\sigma(h_i^T\textbf{S}h_m) \\ 
    P(e=i|X_p)=\sigma(h_i^T\textbf{E}h_m)
\end{aligned}
\end{equation}
where $\textbf{S}$,$\textbf{E}$ are trainable parameter matrices, $h_m$ is the representation of the masked token (i.e., argument mention), $h_i$ is the representation of the token at position $i$, $X_p$ denotes $X_{p_s}$ or $X_{p_j}$. 


For training, we utilize the argument's gold start and end positions $(s_a, e_a)$ to compute the cross-entropy loss for an argument mention:
\begin{equation}
    \mathcal{L}=-\log P(s=s_a|X_p) - \log P(e=e_a|X_p)
\end{equation}

During inference, only the single argument prompt is utilized for argument extraction, since we can not mask the argument mentions in the passage when arguments are unknown yet. Whereas, the templates of the single argument prompt can be adapted from the annotation guidelines.

\section{Experiments}
\subsection{Experimental Setup}
\textbf{Dataset.} 
We evaluate our proposed approach on the most widely used event extraction dataset, namely ACE2005, which contains 599 documents annotated with 33 event subtypes and 35 argument types. For a fair comparison, we follow the same split and preprocessing steps as the previous work \cite{yang-etal-2018-sgm,wadden-etal-2019-entity,du-cardie-2020-event}.

\textbf{Baselines.} 
We compare our approach with the recent advanced baselines\footnote{Given the fact that some argument extraction methods take advantages of entity and relation annotation for extra training tasks, such as \citet{lin-etal-2020-joint} and \citet{zhang-ji-2021-abstract}, we do not include them here.}:
(1) EEQA \cite{du-cardie-2020-event} introduces a new paradigm, i.e., question answering, for argument extraction. 
(2) TANL \cite{tanl} is a encoder-decoder based method using T5, which treats argument extraction as a translation task. (3) BART-GEN \cite{li-etal-2021-conditionalG} is one of the cutting-edge prompt-based methods, which utilizes the manually designed prompt template for argument generation under the encoder-decoder framework.

\textbf{Experimental Settings.} 
The BERT model is used as the backbone. We train our model on a single NVIDIA TiTan XP GPU with Adam as the optimizer. The best learning rate is set as 3e-5 by searching from [2e-5,3e-5,5e-5]. The batch size is optimally set as 32 by selecting among [8,16,32,64]. For other baselines, we follow the hyper parameter settings mentioned in their papers.

\vspace{-2mm}
\subsection{Main Results}
The results are shown in Table \ref{tab:EAE results}, and we have the following observations. \textbf{First}, in both base and large scale settings, our approach surpasses all the baselines with a large magnitude. Our PoKE (large) model achieves the state-of-the-art performance by an absolute improvement of 2.52\% over the best baseline BART-GEN (large) model in terms of F1.
\textbf{Second}, it is interesting to find that our PoKE model in base-scale (PoKE (base)) outperforms the best large-scale baseline, namely (BART-GEN (large)), with an absolute improvement up to 2.16\% regarding F1, which indicates the superiority of our approach in both performance and training efficiency. Overall, compared with the baselines that only use the information of a single event for argument extraction, our approach models the interactions in both intra and inter events with the joint argument prompt, which achieves better performance and higher efficiency.


\begin{table}[t]
\vspace{-2mm}
\small
\caption{Results of event argument extraction.}
\label{tab:EAE results}
\vspace{-2mm}
{\begin{tabular}{c|l|ccc}
\toprule
\textbf{Scale}           & \textbf{Model}    & \textbf{P}     & \textbf{R}     & \textbf{F1}    \\ \midrule
\multicolumn{5}{c}{Main Results} \\
\hline 
\multirow{3}{*}{Base}    & BART-GEN (base)      & 65.04          & 59.94         & 62.39          \\ 
                         & EEQA              & 63.06          & 63.14          & 63.10          \\
                         & PoKE (base)         & \textbf{69.63} & \textbf{68.06}        & \textbf{68.83}          \\ \hline
\multirow{4}{*}{Large}  & TANL    & 65.19          & 64.21          & 64.69          \\ 
                        & EEQA ensemble     & 67.88          & 63.02          & 65.36          \\
                        & BART-GEN (large)     & 67.82    & 65.55          & 66.67          \\
                       & PoKE (large)   & \textbf{71.91} & \textbf{66.67} & \textbf{69.19} \\ \midrule
\multicolumn{5}{c}{Ablation Studies} \\
\hline
\multirow{4}{*}{Base}  
                             & PoKE              & \textbf{69.63} & \textbf{68.06} & \textbf{68.83} \\
                             & - JP  & 67.54          & 67.19          & 67.36          \\
                             & - JP-SP & 63.69          & 61.16          & 62.53          \\ 
                              & - QST & 58.75          & 64.45          & 61.46          \\ \midrule
\multirow{2}{*}{Large}
                             & PoKE              & \textbf{71.91} & \textbf{66.67} & \textbf{69.19} \\ 
                             & PoKE T5           & 68.49           &67.55          & 68.02         \\ \bottomrule     
\end{tabular}}
\vspace{-2mm}
\end{table}

\begin{figure*}[t]
\vspace{-3mm}
     \centering
     \begin{subfigure}{0.3\textwidth}
        \includegraphics[width=\linewidth]{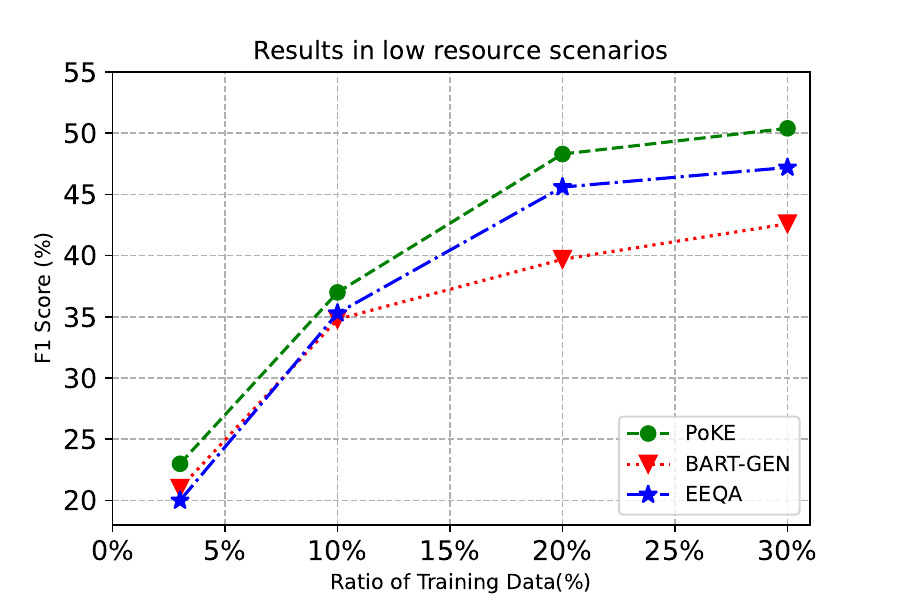}
        \caption{}
        \label{fig:lowResource}
     \end{subfigure}
     \begin{subfigure}{0.3\textwidth}
         \includegraphics[width=\linewidth]{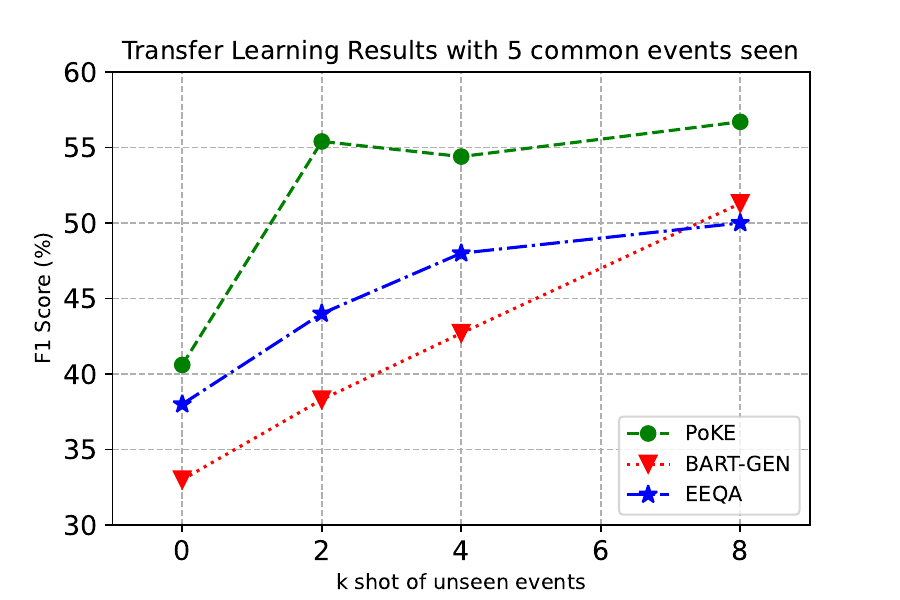}
         \caption{}
         \label{fig: 5 common}
     \end{subfigure}
     \begin{subfigure}{0.3\textwidth}
         \includegraphics[width=\linewidth]{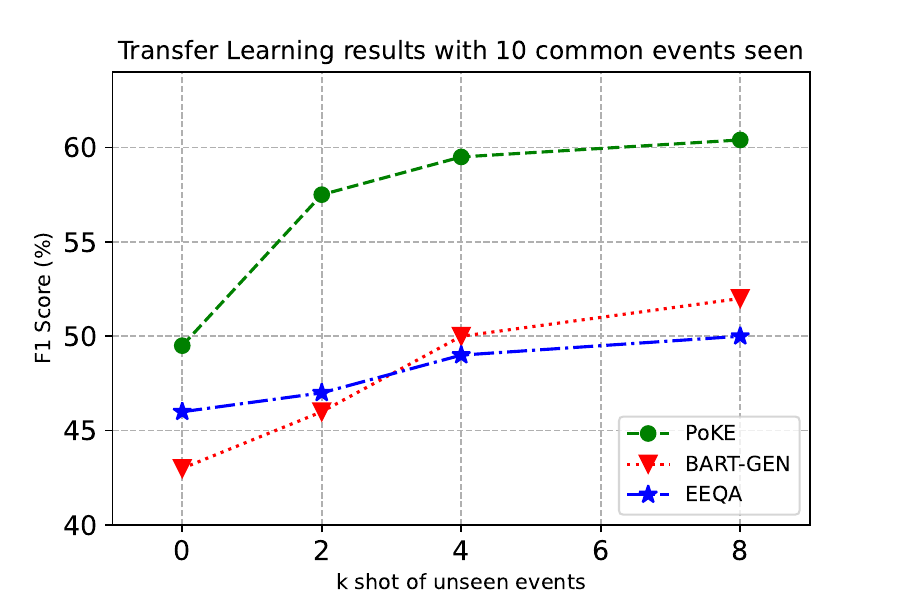}
         \caption{}
         \label{fig: 10 common}
     \end{subfigure}
    \vspace{-3mm}
        \caption{Performance in low resource scenarios and transfer learning setting }
        \label{fig:transferlearning}
    \vspace{-3mm}
\end{figure*}

\subsection{Ablation Studies}
To investigate the effectiveness of each part of our approach, we perform ablation studies. First, we remove the joint argument prompt (-JP) to quantify its effectiveness. Since our approach relies on the single argument prompt in the inference stage, we delete both joint and single argument prompts (-JP-SP) and transform the extraction problem into a QA format to evaluate the effectiveness of both prompts. Besides, we remove the question-aware sequence tagger (-QST) and employ more `[MASK]' tokens to convert the problem into a mask filling paradigm, like \citet{schick-schutze-2021-exploiting}. 
Furthermore, we substitute BERT by T5 (denoted as PoKE T5), which utilizes the encoder-decoder framework to generate the argument instead of using the question-aware sequence tagger.

The results are illustrated in Table \ref{tab:EAE results}. We observe that: (1) The joint argument prompt enhances the performance with approximately 1.5 points in F1 score, which verifies the effectiveness of eliciting more complementary knowledge of intra and inter events from PLMs for argument extraction. 
(2) Without prompts (-JP-SP), the F1 score drops dramatically to 62.53, indicating the great contributions of each prompt strategy for argument extraction.
(3) By replacing the question-aware sequence tagger with mask filling by BERT (-QST) or generation by T5 (PoKE T5), the performance declines significantly, which demonstrates the superiority of our question-aware sequence tagger over other methods.


\subsection{Effectiveness in Low Resource Scenarios}
To verify the effectiveness of our approach in low resource scenarios, we perform experiments with different ratios of training data. 
In particular, we create four datasets covering 3\%, 10\%, 20\% and 30\% of the whole training data respectively. The splits are based on documents, which is more realistic compared to splitting data by instance. Two baselines, namely BART-GEN and EEQA, are used for comparison due to their good performance shown in Table \ref{tab:EAE results}. All the models use the base-scale settings for a fair comparison.

From Figure \ref{fig:lowResource}, we observe that our PoKE model overtakes its counterparts consistently with various ratios of training data. By capturing the argument interactions, our model outperforms the QA-based baseline EEQA by at least 2 points in F1. Regarding the encoder-decoder based baseline, i.e., BART-GEN, its performance falls behind others when the ratio is over 10\%, illustrating the limits of the encoder-decoder framework for argument extraction.

\subsection{Effectiveness in Transfer Learning}
We also investigate the effectiveness of our approach in transfer-learning under zero-shot and few-shot settings. Specifically, we select top $n$ common event types as `seen events' and others as `unseen events', and conduct experiments with $n$=5, 10 respectively. For zero-shot settings, both the training and validation sets are composed of the `seen events' only, while the test set merely contains the `unseen events'. In few-shot settings, we add $k$ samples for each unseen event type into the training and validation sets. We adopt the baselines used in low resource scenarios.

The experiment results are shown in Figure\ref{fig: 5 common} and Figure\ref{fig: 10 common}. It is observed that our model shows superiority over the baselines in both zero-shot and few-shot settings. In zero-shot setting, our PoKE model achieves promising results, surpassing BART-GEN and EEQA with at least 3 points in F1 score. For few-shot setting, we witness a dramatic surge in the performance of our model by merely adding two samples of each unseen event for training, demonstrating the advantage of our model in few-shot learning. Particularly, when $n$=5, our PoKE model trained in 2-shot setting overtakes that trained under zero-shot condition with more than 15\% improvement in F1. In addition, when $n$=10, our model achieves approximately 60\% in F1 by introducing only 4 samples of each unseen event for training, while the baselines EEQA and BART-GEN are struggling around 50\%.

We also observe a slight drop in our performance when $n$=5 and $k$=4. This may be caused by the discrepancy between the distributions of training and testing samples. Such an observation justifies the significance of data sampling, especially in few-shot learning. 
Another interesting observation is that, conditioning on $k$=8, no matter $n$=5 or 10, the performance of EEQA and BART-GEN hardly change. Nonetheless, our PoKE model significantly performs better when $n$=10 than that trained under $n$=5 settings, indicating the strong transfer learning ability with more diverse training data.


\vspace{-0.5em}
\section{Related Work}
Event argument extraction is an important step for event extraction and can be applied in various downstream tasks \cite{NeuralHierarchical,Event-Ranking}.  
Many efforts have been devoted to enhance the performance of EAE, covering most mainstream neural network architectures \cite{wang2019hmeae,ma-etal-2020-resource,xiangyu-etal-2021-BERD}. However, these methods rely heavily on the results of entity recognition.
Recently, there is a trend of modeling event extraction as QA-based tasks \cite{du-cardie-2020-event,liu-etal-2020-event,liu-etal-2021-machine}, getting rid of the dependence on entity annotations. These researches employ reading comprehension abilities of natural language models and rely on large amount of data for training. 
With the popularity of prompt learning \cite{li-liang-2021-prefixtuning,shin-etal-2020-autoprompt,lester-etal-2021-power}, \citet{DBLP:journals/corr/abs-2108-12724,ye2022ontologyenhanced,li-etal-2021-conditionalG} introduce prompt template to elicit knowledge from PLMs for event extraction and achieve promising results. Whereas, the previous methods usually rely on the encoder-decoder architecture, which are prone to error propagation and can not well generate multiple argument mentions when they scatter over the document. In addition, the interactions among arguments and events are not modeled efficiently. 


\section{Conclusions}
We propose a prompt-based knowledge eliciting approach (PoKE), which adapts various prompt strategies for event argument extraction. Extensive experiments have verified the effectiveness of each component of our approach. In particular, our joint argument prompt models the interactions in the intra and inter events, which can help boost the performance by eliciting more complementary knowledge from PLMs. Furthermore, our approach consistently outperforms the strong baselines in both full-training and low-data scenarios. In the future, we will introduce more automatic prompts for joint argument extraction and trigger identification.
%
%

\bibliography{custom}
\bibliographystyle{ACM-Reference-Format}

\appendix

\end{document}